\documentclass{article}
\usepackage{spconf,amsmath,graphicx}

\title{IMAGE EDGE RESTORING FILTER}
\name{Qian Liu, Yongpeng Li, Zhihang Wang}
\address{iQIYI, Beijing, China}

\begin{document}
\maketitle
\begin{abstract}
In computer vision, image processing and computer graphics, 
image smoothing filtering is a very basic and important task and to be expected possessing good edge-preserving smoothing property.
Here we address the problem that the edge-preserving ability of many popular local smoothing filters needs to be improved.
In this paper, we propose the image Edge Restoring Filter (ERF)
to restore the blur edge pixels in the output of local smoothing filters to be clear.
The proposed filter can been implemented after many local smoothing filter (such as Box filter, Gaussian filter, Bilateral Filter, Guided Filter and so on).
The combinations of ``original local smoothing filters + ERF” have better edge-preserving smoothing property than the original local smoothing filters.
Experiments on image smoothing, image denoising and image enhancement demonstrate the excellent edges restoring ability of the proposed filter and good edge-preserving smoothing property of the combination ``original local smoothing filters + ERF”.
The proposed filter would benefit a great variety of applications given that smoothing filtering is a high frequently used and fundamental operation. 
\end{abstract}
\section{Introduction}
\label{sec:intro}

As one of the basic concepts, image filtering plays a very important role in the fields of computer vision, image processing and computer graphics. 
The filter processing is expected to reduce image noise while preserving image edges/structures to the maximum extent. 
In many applications, the results of image filters would largely affect the performances of subsequent tasks. 

\begin{figure}[htb]
  \centerline{\includegraphics[width=9cm]{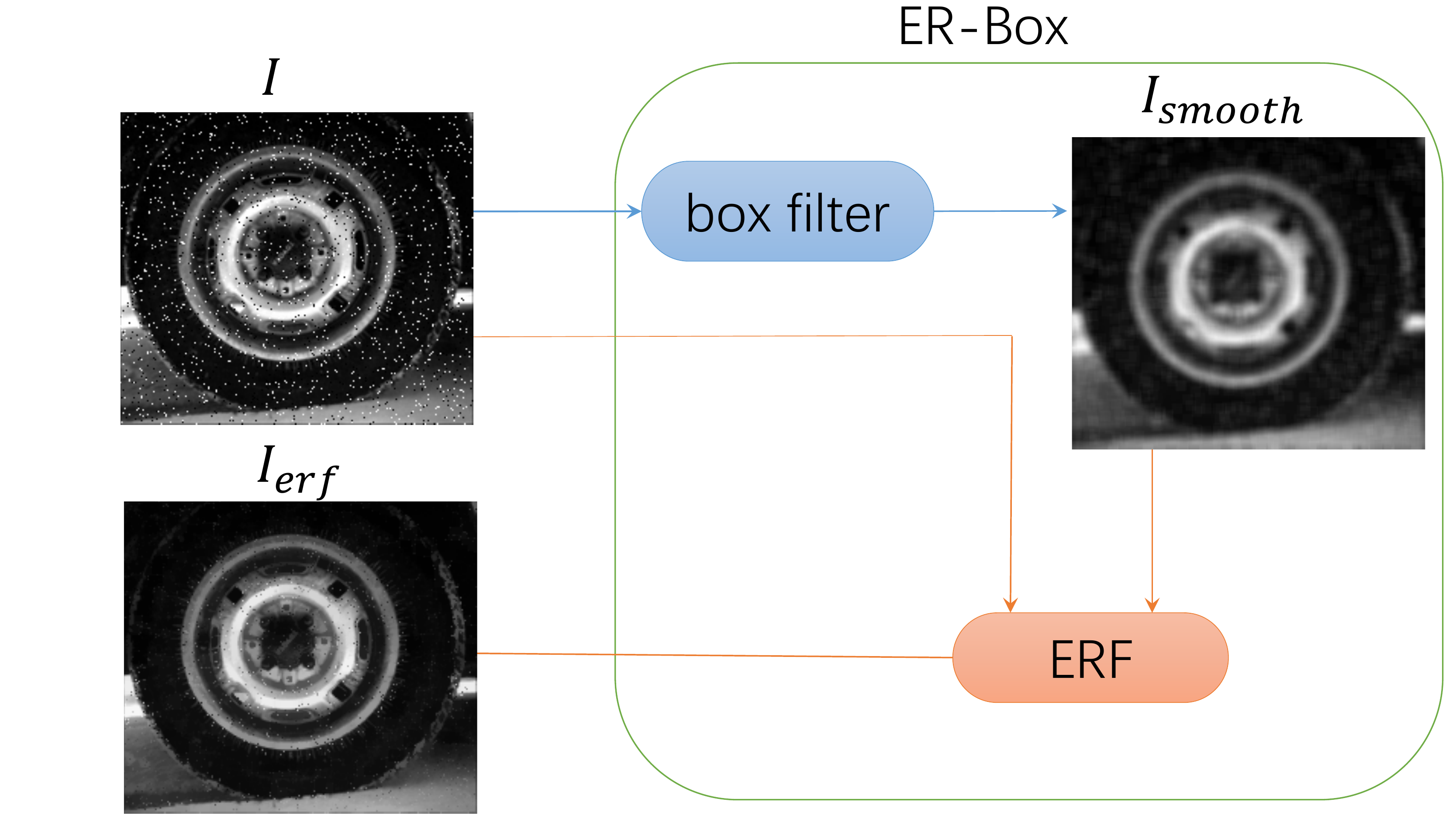}}
\caption{The combinations of ``original local smoothing filters + ERF” (taking ``Box + ERF'' (ER-Box) as an example). The proposed filter ERF restores the blur edges in the smoothing filtering output $I_{smooth}$ to be clear ($r = 3$). } 
\label{f1}
\end{figure}

To obtain good edge preserving capabilities, many methods have been proposed and can be categorized as global filters and local filters. 
Global filters \cite{farbman2008edge} always attempt to compute using information from the whole image. This brings good filtering quality but low computational efficiency.
Local filters utilize the information from the neighborhood of a pixel and achieve lower complexity.
As one of the well-known local filters, the bilateral filter \cite{aurich1995non,tomasi1998bilateral,petschnigg2004digital} is designed to express the pixel value of the filtering output as a weighted average of the given pixel’s neighbor. 
This filter can achieve image denoising while preserving image edges.
The notable guided image filter \cite{he2010guided} yields the output image which having very small intensity difference with the input image and the same edges as the guided image (usually being set as the input image itself for image smoothing/denoising).
The guided image filter and its variants \cite{li2014weighted,ochotorena2019anisotropic} also obtain edge-preserving smoothing results.
However, the edge-preserving property of the above local filters still need to be improved in practical applications.
Center window is always used in traditional local filters. 
In recent techniques, side window filtering \cite{yin2019side} is adopted for better edge-preserving abilities.
In this side window technique \cite{yin2019side}, corresponding to eight specific side windows, the original smoothing filter (e.g. Box filter) is operated eight times. 
It results in increasing calculation amount.
 
Here we consider the phenomenon that edges/structures are blur in the output of many local smoothing filters. Is there some methods having the capability to restore the blur edges in the smoothing filtering output to clear to achieve the better edge-preserving smoothing results?

In this paper, we propose the image Edge Restoring Filter (ERF) to address this problem.
The ERF is designed as a post process filter and is implemented after the original local smoothing filter, such as Box filter (Box), Gaussian filter (Gau), Bilateral filter (Bil), Guided filter (Gui) and so on. 
Considering the input image $I$ and its local smoothing filtered output $I_{smooth}$, which normally presents blur edges after local smoothing filtering, the proposed ERF operates on a pair of images ($I_{smooth}$ and $I$) and aims to restore blur edges of $ I_{smooth}$ to be clear.  
The combination of “local smoothing filter + ERF” can better achieve edge-preserving smoothing property, 
since ERF has ability to restore blur edges in $ I_{smooth}$ to clear.
The ERF is inspired by side window filtering \cite{yin2019side}. The ERF is more convenient to be applied and bring better edge-preserving property than \cite{yin2019side}.

Figure 1 is an example of ``Box + ERF'' (ER-Box). As shown in Figure \ref{f1}, $I_{smooth}$ is the smoothing result of $I$ and has very blur edges. 
ERF is used to restore the blur edges in $I_{smooth}$ to be clear and outputs the edge restoring result $I_{erf}$. 
As the output of ER-Box, $I_{erf}$ achieves better edge-preserving smoothing result than $I_{smooth}$.  

\section{Image edge restoring filter}
\label{sec:erf}

\subsection{Definition}
\label{sec:erf-1}

We firstly define the proposed image edge restoring filter (ERF). 

Consider the input image $I$ and its corresponding smoothing filtering output $I_{smooth}$ which is filtered on the center filtering windows with radius $r$:
\begin{equation}\label{eq0}
 I_{smooth}  = Smooth(I, r),
\end{equation}
where $Smooth()$ denotes one local smoothing filter, 
such as Box, Gau, Bil, Gui and so on.
The proposed image edge restoring filter (ERF) is a post process filter and is implemented after Eq \ref{eq0}. 
ERF generates the result $I_{erf}$ via operating on a pair of images $I_{smooth}$ and $I$:
\begin{equation}\label{eq1}
 I_{erf}  = ERF(I_{smooth}, I, r),
\end{equation}
where $ERF()$ is the proposed filter and $r$ is the filter radius. 
Note that the filtering radius $r$ in Eq \ref{eq1} is strictly the same as that in Eq \ref{eq0} (the original smoothing filtering).
Let's consider a target pixel with index $i$. 
If pixel $I(i)$ belongs to edge pixels and $I_{smooth}(i)$ is blur, it is expected that $I_{erf}(i)$ is close to $I(i)$.
Oppositely, it is not expected that $I_{erf}(i)$ is equal to $I(i)$ if $I(i)$ belongs to noise pixels.
In order to achieve this goal, we design ERF as follows. 
Find pixel $j'$ in $I_{smooth}$ by sloving:
\begin{equation}\label{eq2}
{\bf{argmin}}_{j \in\omega_i} D(I_{smooth}(j), I(i)),
\end{equation}
where $\omega_i$ is a window centered at the pixel $i$ in image $I_{smooth}$, 
$D(I_{smooth}(j), I(i))$ is $(I_{smooth}(j)-I(i))^2$ for gray image and $(I_{smooth}(j)^r-I(i)^r)^2 + ( I_{smooth}(j)^g-I(i)^g)^2 + (I_{smooth}(j)^b-I(i)^b)^2$ for color image.
After obtaining the value of $j'$, we get $I_{erf}(i)$:
\begin{equation}\label{eq4}
I_{erf}(i) = I_{smooth}(j').
\end{equation}

\subsection{Implement}
\label{sec:erf-3}
It is very convenient to equip many popular local smoothing filters (such as Box, Gau, Bil, Gui and so on) with the proposed filter ERF to obtain better edge-preserving smoothing results.  
Taking ``Box + ERF'' (ER-Box) as an instance, let us give the overall stages of ER-Box.
As shown in Algorithm 1, ER-Box consists of two steps: implementing Box filter, then implementing ERF filter. 
Since the computation for each pixel $i$ in $I_{erf}$ is individual, parallel computing can be applied to accelerate ERF.

To generalize to more filters, Box filter in Algorithm 1 can be directly replaced by Gau, Bil, Gui or other local smoothing filter to get ER-Gau,  ER-Bil, ER-Gui or others. 

\begin{table}    
\label{t1}
\begin{center}  
\begin{tabular}{l}  
\hline  
\bf{Algorithm 1} ER-Box  \\
\hline 
{\bf{Input}}: $I$ and $r$ \\
$I$ is the original image, and $r$ is the window radius  used \\
 in both Box and ERF. \\
{\bf{Output}}: the result image $I_{erf}$ \\
1: computing $I_{smooth}$ \\
$I_{smooth} = {\bf{boxfilter}}(I, r)$  \\
2: for each pixel $i$ in image $I_{erf}$: finding pixel $j'$ by \\
sloving Eq. \ref{eq2} and computing $I_{erf}(i)$ via Eq. \ref{eq4}. \\
Output: $I_{erf}$ \\
\hline  
\end{tabular}  
\end{center}  
\end{table}

 \begin{figure*}[hbt]
\begin{minipage}[b]{0.16\linewidth}
  \centering
  \centerline{\includegraphics[width=2.8cm]{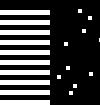}}
  \centerline{(a) input}\medskip
\end{minipage}
\begin{minipage}[b]{.16\linewidth}
  \centering
  \centerline{\includegraphics[width=2.8cm]{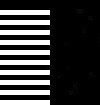}}
  \centerline{(b) ER-Box}\medskip
\end{minipage}
\begin{minipage}[b]{.16\linewidth}
  \centering
  \centerline{\includegraphics[width=2.8cm]{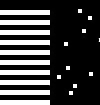}}
  \centerline{(c) S-Box(r=1)}\medskip
\end{minipage}
\begin{minipage}[b]{.16\linewidth}
  \centering
  \centerline{\includegraphics[width=2.8cm]{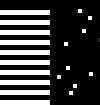}}
  \centerline{(d)S-Box(r=2)}\medskip
\end{minipage}
\begin{minipage}[b]{.16\linewidth}
  \centering
  \centerline{\includegraphics[width=2.8cm]{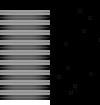}}
  \centerline{(e) S-Box(r=3)}\medskip
\end{minipage}
\begin{minipage}[b]{.16\linewidth}
  \centering
  \centerline{\includegraphics[width=2.8cm]{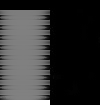}}
  \centerline{(f) S-Box(r=4)}\medskip
\end{minipage}
\caption{Edge-preserving smoothing.}
\label{fig:3-0}
\end{figure*}

\begin{figure*}[htb]

\begin{minipage}[b]{0.19\linewidth}
  \centering
  \centerline{\includegraphics[width=3.4cm]{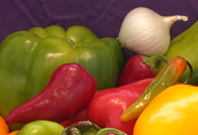}}
  \centerline{(a) input}\medskip
\end{minipage}
\begin{minipage}[b]{.19\linewidth}
  \centering
  \centerline{\includegraphics[width=3.4cm]{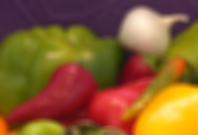}}
  \centerline{(b) Box}\medskip
\end{minipage}
\begin{minipage}[b]{.19\linewidth}
  \centering
  \centerline{\includegraphics[width=3.4cm]{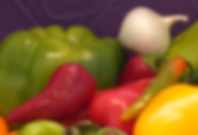}}
  \centerline{(c) Gau}\medskip
\end{minipage}
\begin{minipage}[b]{.19\linewidth}
  \centering
  \centerline{\includegraphics[width=3.4cm]{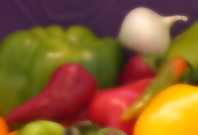}}
  \centerline{(d) Bil}\medskip
\end{minipage}
\begin{minipage}[b]{.19\linewidth}
  \centering
  \centerline{\includegraphics[width=3.4cm]{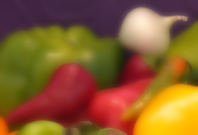}}
  \centerline{(e) Gui}\medskip
\end{minipage}

 \hspace{3.4cm}
\begin{minipage}[b]{0.19\linewidth}
  \centering
  \centerline{\includegraphics[width=3.4cm]{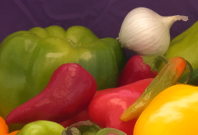}}
  \centerline{(f) ER-Box}\medskip
\end{minipage}
\begin{minipage}[b]{.19\linewidth}
  \centering
  \centerline{\includegraphics[width=3.4cm]{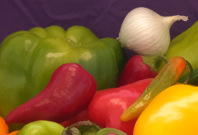}}
  \centerline{(g) ER-Gau}\medskip
\end{minipage}
\begin{minipage}[b]{.19\linewidth}
  \centering
  \centerline{\includegraphics[width=3.4cm]{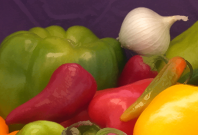}}
  \centerline{(h) ER-Bil}\medskip
\end{minipage}
\begin{minipage}[b]{.19\linewidth}
  \centering
  \centerline{\includegraphics[width=3.4cm]{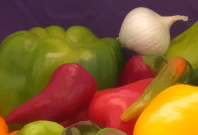}}
  \centerline{(i) ER-Gui}\medskip
\end{minipage}
\caption{Image smoothing.}
\label{fig:3-1}
\end{figure*}

\subsection{Analysis}
\label{sec:erf-2}
The proposed ERF can achieve the goal that restoring the edge pixels from the smoothing filtered image.
The work mechanism can be understood as follows.    

Taking ER-Box as an example, consider two cases.

Case 1: ``edge pixel''. If pixel $i$ belongs to the edge/structure pixels in $I$, it is observed that there is often one pixel $j'$ in the neighborhood of pixel $i$ in $I_{smooth}$ to meet the condition that the value $I_{smooth}(j')$ is close or equal to $I(i)$. So, the output value of Eq \ref{eq4} can restore $I_{smooth}(i)$ back to $I(i)$. 

Case 2: ``noise pixel''. If pixel $i$ is noise in $I$ and pixel $j$ is in the neighborhood of pixel $i$ in $I_{smooth}$, the value $I_{smooth}(j)$ which is close to $I(i)$ often does not exist. 

As a result, the proposed filter can distinguish edge/structure pixels from noise pixels and ERF can restore blur edges pixels in $I_{smooth}$ to be clear.

Furthermore, ERF uses the information of multiple smoothing filtering results of pixel $i$, which includes side window smoothing filtering results, eccentric window (the target pixel is inside the filter window, but not at the center or on the side) smoothing filtering results and center window smoothing filtering results.  
In fact, in $I$, if $j \in \omega_i$ and $j \neq  i$, the center filtering window of pixel $j$ is right one side window or eccentric window of pixel $i$. So, $I_{smooth}(j)$, which is the the center window filtering result of pixel $j$, is right the side filtering window or eccentric filtering window smoothing filtering result of pixel $i$.

Compared to \cite{yin2019side}, ERF is more convenient, use more information and expects better results. 
In \cite{yin2019side}, corresponding to eight specific side windows (down, right, up, left, southwest, southeast, northeast and northwest side windows), the original smoothing filter (e.g. Box) is implemented eight times.    
As a contrast, the original smoothing filter is implemented only once and normally in the center window in ERF.
Our ERF uses more information from side window results, eccentric window results and center window result, compared to \cite{yin2019side} which utilizes only few side windows results.

\section{Experiment}
\label{sec:exp}
To demonstrate the edge restoring ability of the proposed ERF and the edge preserving smoothing property of “local smoothing filter + ERF”, we evaluate ERF on some challenge cases in \ref{sec:Edge-preserving} and several applications in \ref{sec:application}: 
image smoothing, image denoising and image enhancement. 
All codes are implemented in Matlab. 
Box/Gau/Bil/Gui are implemented via functions provided in Matlab R2020b. 
These results images are best viewed in color and electronically.

\begin{figure*}[htb]

\begin{minipage}[b]{0.19\linewidth}
  \centering
  \centerline{\includegraphics[width=3.4cm]{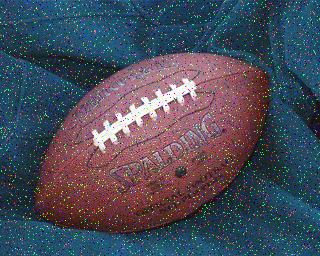}}
  \centerline{(a) input}\medskip
\end{minipage}
\begin{minipage}[b]{.19\linewidth}
  \centering
  \centerline{\includegraphics[width=3.4cm]{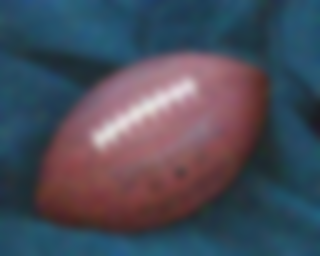}}
  \centerline{(b) Box}\medskip
\end{minipage}
\begin{minipage}[b]{.19\linewidth}
  \centering
  \centerline{\includegraphics[width=3.4cm]{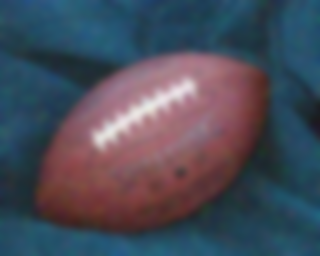}}
  \centerline{(c) Gau}\medskip
\end{minipage}
\begin{minipage}[b]{.19\linewidth}
  \centering
  \centerline{\includegraphics[width=3.4cm]{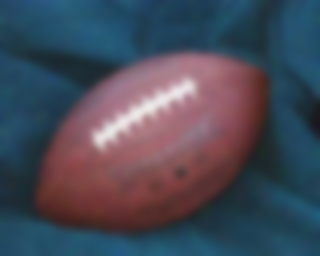}}
  \centerline{(d) Bil}\medskip
\end{minipage}
\begin{minipage}[b]{.19\linewidth}
  \centering
  \centerline{\includegraphics[width=3.4cm]{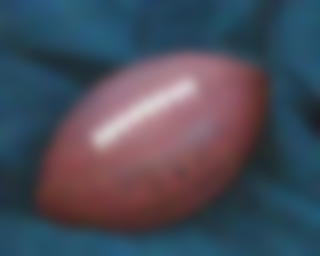}}
  \centerline{(e) Gui}\medskip
\end{minipage}

 \hspace{3.4cm}
\begin{minipage}[b]{0.19\linewidth}
  \centering
  \centerline{\includegraphics[width=3.4cm]{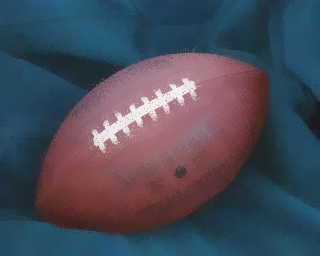}}
  \centerline{(f) ER-Box}\medskip
\end{minipage}
\begin{minipage}[b]{.19\linewidth}
  \centering
  \centerline{\includegraphics[width=3.4cm]{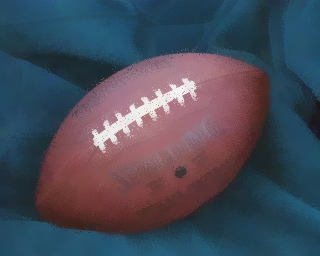}}
  \centerline{(g) ER-Gau}\medskip
\end{minipage}
\begin{minipage}[b]{.19\linewidth}
  \centering
  \centerline{\includegraphics[width=3.4cm]{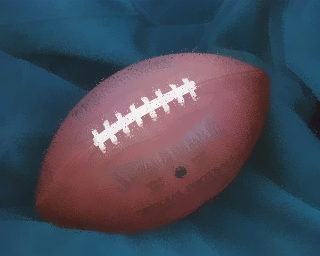}}
  \centerline{(h) ER-Bil}\medskip
\end{minipage}
\begin{minipage}[b]{.19\linewidth}
  \centering
  \centerline{\includegraphics[width=3.4cm]{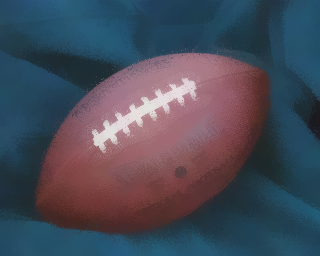}}
  \centerline{(i) ER-Gui}\medskip
\end{minipage}
\caption{Image denoising.}
\label{fig:3-2}
\end{figure*}

\begin{figure*}[htb]
\begin{minipage}[b]{0.19\linewidth}
  \centering
  \centerline{\includegraphics[width=3.4cm]{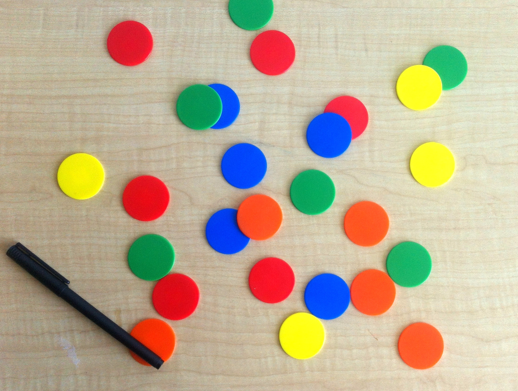}}
  \centerline{(a) input}\medskip
\end{minipage}
\begin{minipage}[b]{.19\linewidth}
  \centering
  \centerline{\includegraphics[width=3.4cm]{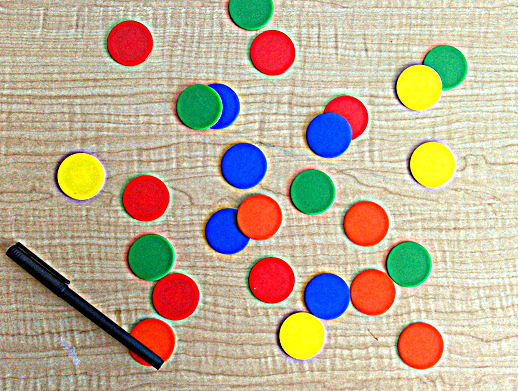}}
  \centerline{(d) Bil}\medskip
\end{minipage}
\begin{minipage}[b]{.19\linewidth}
  \centering
  \centerline{\includegraphics[width=3.4cm]{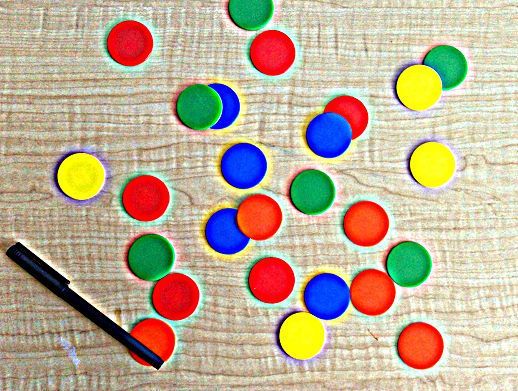}}
  \centerline{(e) Gui}\medskip
\end{minipage}
%
\begin{minipage}[b]{.19\linewidth}
  \centering
  \centerline{\includegraphics[width=3.4cm]{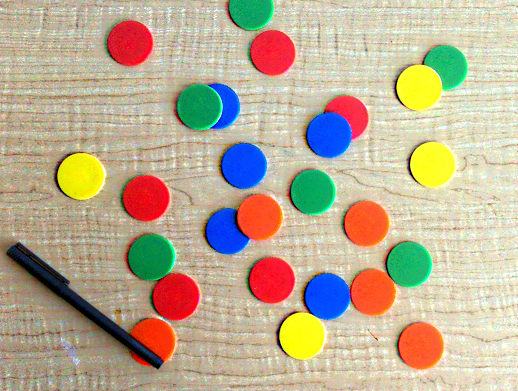}}
  \centerline{(h) ER-Bil}\medskip
\end{minipage}
\begin{minipage}[b]{.19\linewidth}
  \centering
  \centerline{\includegraphics[width=3.4cm]{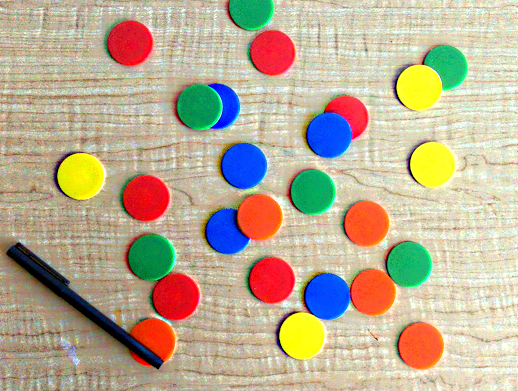}}
  \centerline{(i) ER-Gui}\medskip
\end{minipage}
\caption{Image denoising.}
\label{fig:3-3}
\end{figure*}

\subsection{Edge-preserving smoothing property}
\label{sec:Edge-preserving}

The edge-preserving smoothing property of ERF is demonstrated in Figure \ref{fig:3-0}.
Here we focus on some extreme and challenging cases, e.g. the size of edges and the size of noise patches are very close.
In the input image Figure \ref{fig:3-0}(a), edges are on left part with a width 5 pixels and right part distributes some noise patches with a width 4 pixels.
Ideally, all of noise patches are removed and all of the edge pixels are preserved after smoothing filtering.
We compared ER-Box with S-Box (side windows filter + Box filter) \cite{yin2019side} here.
For \cite{yin2019side}, author-provided codes are used and downloaded from Github.
In Figure \ref{fig:3-0}, all results are obtained after 10 iterations.
Figure \ref{fig:3-0}(b) is the result of ER-Box. It preserves the edges of Figure \ref{fig:3-0}(a) and smooths the noise pixels thoroughly. 
The results of S-Box can not preserve the edges when $r=3$ and $r=4$. For $r=1$ and $r=2$, S-Box can not remove the noise patches.
The observation is obtained from Figure \ref{fig:3-0} that ER-Box gives better edge preserving smoothing results compared to S-Box.

\subsection{Application}
\label{sec:application}

{\bf{Image smoothing.}}
We compare ER-Box/ER-Gau/ER-Bil/ER-Gui and their original counterparts Box/Gau/Bil/Gui for image smoothing.
The results are shown in Figure \ref{fig:3-1}. 
The parameters are set as follows: $r =3$ for Box/ER-Box, $r=3, \sigma = 2$ for Gau/ER-Gau,  $r=3$, $\sigma_s = 3$, DegreeOfSmoothing ($dos$) = 0.3 for Bil/ER-Bil, $r=3$, $dos$ = 0.1 for Gui/ER-Gui. 
These original filters provide the results with the blur edges. As a comparison, ER-Box/ER-Gau/ER-Bil/ER-Gui achieve the clearer edges. 
This demonstrates excellent edge restoring ability of the proposed ERF.

{\bf{Image denoising.}}
Figure \ref{fig:3-2} shows the results of image denoising. 
The results are obtained by iteratively applying the filters for 5 times.
The parameters are set as the same as those used in Figure \ref{fig:3-1}.
Contrary to the blur edges in outputs of the original filters, clearer and better edges in the outputs of ER-Box/ER-Gau/ER-Bil/ER-Gui demonstrate the excellent edge restoring ability of the proposed filter ERF.

{\bf{Image enhancement.}}
The results of image enhancement are presented in Figure \ref{fig:3-3}.
The enhanced images are obtained by 
\begin{equation}
I_{en}=I_{base}+C*(I–I_{base}),
\end{equation}
where, $I$ is the original input image, $I_{base}$ is the output of Bil/Gui/ER-Bil/ER-Gui, and $C$ is the amplification parameter and is set as 5. 
In Figure \ref{fig:3-3}, the parameters are set as follows: $r=5$, $\sigma_s = 5$, $dos$ = 0.3 for Bil/ER-Bil, $r=5$, $dos$ = 0.1 for Gui/ER-Gui.
The results of Bil/Gui cause edge halos along the color circular chips. Whereas, these edge halos are improved and avoided in the results of ER-Bil/ER-Gui. From Figure \ref{fig:3-3}, you can find that good edge preserving property of ER-Bil/ER-Gui brings uniformly enhancement.

\section{Conclusion}
\label{sec:con}
In this paper, the image edge restoring filter (ERF) is proposed.
ERF is designed as a post process filter and has the ability of restoring the blur edges of local smoothing filtering output to be clearer. 
It is very convenient to use ERF after many popular local smoothing filters. 
The combination of ``original local smoothing filters + ERF'' has better edge-preserving smoothing abilities than the original filters.
Experiments results show the excellent edges restoring ability of ERF and the good edge-preserving smoothing property of the combination of ``original local smoothing filters + ERF''.

\bibliographystyle{IEEEbib}
\bibliography{erf}

\end{document}